\pdfoutput=1
\documentclass[english,runningheads,a4paper]{llncs}
\usepackage[english]{babel}
\addto\extrasenglish{\languageshorthands{english}\useshorthands{"}}

\usepackage{regexpatch}
\makeatletter
\edef\switcht@albion{%
  \relax\unexpanded\expandafter{\switcht@albion}%
}
\xpatchcmd*{\switcht@albion}{ \def}{\def}{}{}
\xpatchcmd{\switcht@albion}{\relax}{}{}{}
\edef\switcht@deutsch{%
  \relax\unexpanded\expandafter{\switcht@deutsch}%
}
\xpatchcmd*{\switcht@deutsch}{ \def}{\def}{}{}
\xpatchcmd{\switcht@deutsch}{\relax}{}{}{}
\edef\switcht@francais{%
  \relax\unexpanded\expandafter{\switcht@francais}%
}
\xpatchcmd*{\switcht@francais}{ \def}{\def}{}{}
\xpatchcmd{\switcht@francais}{\relax}{}{}{}
\makeatother

\usepackage{ifluatex}
\ifluatex
  \usepackage{fontspec}
  \usepackage[english]{selnolig}
\fi

\iftrue % use default-font
  \ifluatex
  \else
    \usepackage[%
      rm={oldstyle=false,proportional=true},%
      sf={oldstyle=false,proportional=true},%
      tt={oldstyle=false,proportional=true,variable=false},%
      qt=false%
    ]{cfr-lm}
  \fi
\else
  \ifluatex
  \else
    \usepackage{newtxtext}
    \usepackage{newtxmath}
    \usepackage[zerostyle=b,scaled=.9]{newtxtt}
  \fi
\fi

\ifluatex
\else
  \usepackage[T1]{fontenc}
  \usepackage[utf8]{inputenc} %support umlauts in the input
\fi

\usepackage{graphicx}
\usepackage{upquote}
\usepackage{booktabs}
\usepackage{paralist}
\usepackage{csquotes}
\usepackage{textcmds}

\RequirePackage[%
  babel,%
  final,%
  expansion=alltext,%
  protrusion=alltext-nott]{microtype}%
\DisableLigatures{encoding = T1, family = tt* }

\usepackage{url}
\makeatletter
\g@addto@macro{\UrlBreaks}{\UrlOrds}
\makeatother
\usepackage{xcolor}

\usepackage{listings}
\renewcommand{\lstlistingname}{List.}
\usepackage{chngcntr}
\AtBeginDocument{\counterwithout{lstlisting}{section}}

\usepackage{pdfcomment}
\usepackage[%
  square,        % for square brackets
  comma,         % use commas as separators
  numbers,       % for numerical citations;
  sort&compress, % as sort but in addition multiple numerical citations
]{natbib}
\ifluatex
\else
  \SetExpansion
  [ context = sloppy,
    stretch = 30,
    shrink = 60,
    step = 5 ]
  { encoding = {OT1,T1,TS1} }
  { }
\fi

\usepackage{stfloats}
\fnbelowfloat
\usepackage{hyperref}
\usepackage[all]{hypcap}

\usepackage[group-four-digits,per-mode=fraction]{siunitx}

\usepackage[capitalise,nameinlink]{cleveref}
\usepackage{iflang}
\IfLanguageName{ngerman}{
  \crefname{table}{Tab.}{Tab.}
  \Crefname{table}{Tabelle}{Tabellen}
  \crefname{figure}{\figurename}{\figurename}
  \Crefname{figure}{Abbildungen}{Abbildungen}
  \crefname{equation}{Gleichung}{Gleichungen}
  \Crefname{equation}{Gleichung}{Gleichungen}
  \crefname{listing}{\lstlistingname}{\lstlistingname}
  \Crefname{listing}{Listing}{Listings}
  \crefname{section}{Abschnitt}{Abschnitte}
  \Crefname{section}{Abschnitt}{Abschnitte}
  \crefname{paragraph}{Abschnitt}{Abschnitte}
  \Crefname{paragraph}{Abschnitt}{Abschnitte}
  \crefname{subparagraph}{Abschnitt}{Abschnitte}
  \Crefname{subparagraph}{Abschnitt}{Abschnitte}
}{
  \crefname{section}{Sect.}{Sect.}
  \Crefname{section}{Section}{Sections}
  \crefname{listing}{\lstlistingname}{\lstlistingname}
  \Crefname{listing}{Listing}{Listings}
}

\usepackage{xspace}

\DeclareFontFamily{U}{MnSymbolC}{}
\DeclareSymbolFont{MnSyC}{U}{MnSymbolC}{m}{n}
\DeclareFontShape{U}{MnSymbolC}{m}{n}{
  <-6>    MnSymbolC5
  <6-7>   MnSymbolC6
  <7-8>   MnSymbolC7
  <8-9>   MnSymbolC8
  <9-10>  MnSymbolC9
  <10-12> MnSymbolC10
  <12->   MnSymbolC12%
}{}
\DeclareMathSymbol{\powerset}{\mathord}{MnSyC}{180}

\ifluatex
\else
  \input glyphtounicode
\fi
\usepackage[math]{blindtext}
\usepackage{mwe}

\usepackage{graphicx}

\usepackage{perpage} %the perpage package
\MakePerPage{footnote} %the perpage package command

\begin{document}
\title{An Intelligent Safety System for Human-Centered Semi-Autonomous Vehicles}
\titlerunning{An Intelligent Safety System for Semi-Autonomous Vehicles}
\author{Hadi {Abdi Khojasteh}\inst{1} \and
Alireza {Abbas Alipour}\inst{1} \and
Ebrahim Ansari\inst{1,2}\\ \and
Parvin Razzaghi\inst{1,3}}
\authorrunning{H. {Abdi Khojasteh} et al.}
\institute{Department of Computer Science and Information Technology,\\Institute for Advanced Studies in Basic Sciences (IASBS), Zanjan, Iran\\
\email{\{hkhojasteh,alr.alipour,ansari,p.razzaghi\}@iasbs.ac.ir}\\
\url{https://iasbs.ac.ir/\textasciitilde ansari/faraz} \and
Charles University, Faculty of Mathematics and Physics,\\Institute of Formal and Applied Linguistics \and
School of Computer Science,\\Institute for Research in Fundamental Sciences (IPM), Tehran, Iran}
\maketitle
\begin{abstract}
Nowadays, automobile manufacturers make efforts to develop ways to make cars fully safe. Monitoring driver's actions by computer vision techniques to detect driving mistakes in real-time and then planning for autonomous driving to avoid vehicle collisions is one of the most important issues that has been investigated in the machine vision and Intelligent Transportation Systems (ITS). The main goal of this study is to prevent accidents caused by fatigue, drowsiness, and driver distraction. To avoid these incidents, this paper proposes an integrated safety system that continuously monitors the driver's attention and vehicle surroundings, and finally decides whether the actual steering control status is safe or not. For this purpose, we equipped an ordinary car called FARAZ with a vision system consisting of four mounted cameras along with a universal car tool for communicating with surrounding factory-installed sensors and other car systems, and sending commands to actuators. The proposed system leverages a scene understanding pipeline using deep convolutional encoder-decoder networks and a driver state detection pipeline. We have been identifying and assessing domestic capabilities for the development of technologies specifically of the ordinary vehicles in order to manufacture smart cars and eke providing an intelligent system to increase safety and to assist the driver in various conditions/situations.

\end{abstract}
\section{Introduction}
According to the World Health Organization (WHO) in 2013, some 1.4 million people lose their lives in traffic accidents each year~\cite{world2013global}. Also, a 2009 report published by the WHO had estimated that more than 1.2 million people die and up to 50 million people are injured or disabled in road traffic crashes around the world every year~\cite{world2009global}. The statistics show that, due to the ever-increasing number of vehicles and density of traffic on roads, current intelligent transportation systems have been successful. However, the systems need to be further developed to decrease the number and severity of road accidents.

The Integrated Vehicle Safety System (IVSS)~\cite{green2008integrated} is used for safety applications in vehicles. The system which includes various safety systems such as anti-lock braking system (ABS), emergency brake assist (EBS), traction control system (known as ASR), crash mitigation systems, and lane keeping assist systems. The purpose of an IVSS is to provide all safety related functions for all types of vehicles at a minimum cost. Such system offers several advantages like low cost, compact size, driving comfort, traffic information, and safety alerts. It also indicates the health of the car electrical components and provides information regarding an overall condition of the vehicle.

\begin{figure}
	\centering
	\includegraphics[width=0.98\textwidth]{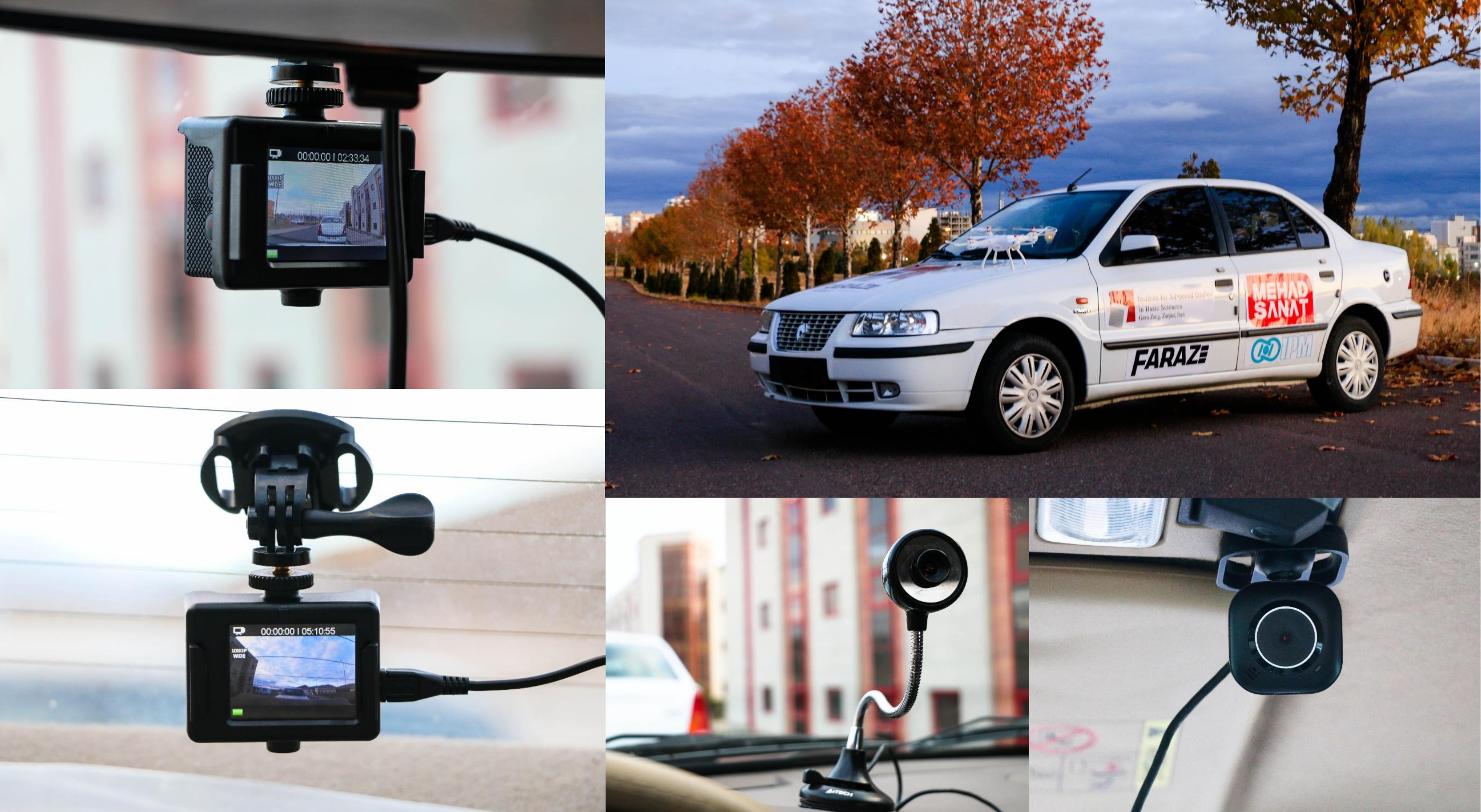}
	\caption{The instrumented vehicle and drone (top right) with the vision system consists of four mounted cameras and a drone camera along with a universal car tool for communicating and sending commands to the vehicle. The front (top left) and rear (bottom left) wide-angle HD cameras are mounted at close to the center of the windshields. The driver-facing camera (bottom left) is mounted on the center of the roadway view. The car cabin camera (bottom right) is mounted on the center of the headliner to include a view of the driver's body.}
	\label{fig:instrumented-vehicle}
\end{figure}
In the past decade, many studies have examined the advantages of integrated safety and driver acceptance along with integrated crash warning systems. A study was undertaken by the U.S. Department of Transportation indicates that number of crashes can be reduced significantly by developing collision warning systems to alert drivers of potential rear-end, lateral drift, and lane-changing crashes~\cite{green2008integrated}. Such integrated warning systems will provide comprehensive and coordinated information, which can be used by crash warning subsystems to warn the driver of the foreseeable threats. For an intelligent system to reduce accidents and casualties, at least two general trends can be expected: \textit{1) Autonomous cars, 2) Driver assistance systems.}

An autonomous car also known as a self-driving car is a vehicle that has the characteristics of a traditional car and in addition, is capable of transporting automatically without human intervention. A driverless car system drives the vehicle by the perception of the environment and based on dynamic processes which result in steering and navigating the car to safety~\cite{bojarski2016end,hee2013motion}. As it seems, the studies done to get to self-driving cars have led to creating the driver assistance systems. From another perspective, an utter vehicle control system, without examining different driver assistance systems, as well as the use of intelligent highways, is meaningless. Design and implementation of automated driving in the real environment, with regards to today's technology is still in the preliminary stages~\cite{cheng2007interactive} and there is a long road to its full implementation. As a short-term and practical solution, today, much effort is being made in the research and industrial communities to design and implement driver assistance systems. For example, the automation of some areas of vehicle control, such as auto-steering, or the movement direction, has been widely studied and implemented or studies carried out on various driving maneuvers such as overtaking and automatic parking of the vehicle~\cite{naranjo2008lane}.

Deep learning~\cite{innocenti2017imitation,neven2018towards,vicente2015driver} can be used to analyze and process input data received from the automobile sensors such as cameras, motion sensors, laser light, GPS, Odometry, LiDAR, and radar sensors and control the vehicle in response to information from various sensors~\cite{fridman2016owl,hoffman2000distinct,wisniewska2014robust,bojarski2016end,hee2013motion}. Also, we can utilize computer vision for eye-gaze tracking~\cite{fridman2016owl,hoffman2000distinct,wisniewska2014robust,vicente2015driver}, monitoring threshold blinking~\cite{soukupova2016real} and head movement~\cite{varma2012accident} which in turn such in-car sensing technologies would enable us to warn the driver of drowsiness or distraction in real-time. Hence, these measures can be highly effective in avoiding collisions and reducing fatal accidents.

Our central goal in this work is to create a semi-autonomous car by integrating some state-of-the-art approaches in computer vision and machine learning for assisting the drivers during critical and risky moments in which driver would be unable to steer the vehicle safely. All supplementary materials are available for public access on the web\footnote{\url{https://iasbs.ac.ir/\textasciitilde ansari/faraz}}.

The rest of the paper is organized as follows. We review former state-of-the-art approaches in \Cref{sec:Background}. In \Cref{sec:Architecture}, we describe our safety system architecture on the semi-autonomous car  in which we applied some subtle manipulations along with ordinary capabilities of traditional vehicles. \Cref{sec:Implementation-Details} explains our fine-grained system in detail. The paper concludes with \Cref{sec:Conclusions-and-Future-Work} where we discuss the outcome and possible work to be done in the future.

\section{Background}\label{sec:Background}
There are many works on preventing car accidents some of which deal with the effects of driver behavior in traffic accidents. As in~\cite{moghaddam2014introducing} authors have carried out research on the drive in which they use the raw data that is collected for processing to define driving violations as a criterion for driving behavior and have examined the impact of various factors such as speed,  the effect of density, velocity, and traffic flow on accidents. Much research has introduced automotive safety systems which designed to avoid or reduce the severity of the collision. In such collision mitigating systems, tools like radar, laser (LiDAR) and camera (employing image recognition) are utilized to detect an imminent crash~\cite{choi2012environment}. Many articles have been presented to prevent crashes with the use of intelligent systems. Some systems react to imminent crash (occurring at the moment). As an example, in~\cite{cheng2007interactive}, using parameters like speed and distance of vehicles, the systems help prevent collisions at intersections or reduce damage and casualties. Some consider the current condition of the road and neighboring cars and using the available data, examine the probability of an accident and predict them to provide solutions to avoid accidents.

Moreover, an early work proposed a traffic-aware cruise control system for road transports that automatically tunes the vehicle speed to keep an assured distance from other cars ahead. Such systems might utilize various sensors such as radar, LiDAR, or a stereo camera system for the vehicle to brake when the system finds the vehicle is approaching another car ahead, then accelerate when traffic allows it to. One of the most common types of accidents is rear-end crashes which accounts for a significant percentage of accidents in different countries~\cite{naranjo2008lane}. Rate of these accidents are even more frequent on the roads. In order to avoid rear-end accidents two solutions are considered: timely change of speed which is when the vehicle detects that a collision with the front (rear) vehicle is imminent, the speed is reduced/increased to prevent it, and changing the direction in order to prevent collisions with the front or rear car, the driver changes the car's path.

Most of the research focused on vision-based methods which used to assist the driver for steering a vehicle safely and comfortably. In~\cite{fridman2018human} authors proposed an approach in which they use only cameras and machine learning techniques to perform the driving scene perception, motion planning, driver sensing to implement the seven principles that they described in the work for making a human-centered autonomous vehicle system~\cite{fridman2018human}. Also author in~\cite{alessandretti2007vehicle} fused radar and camera data to improve the perception of the vehicle's surrounding, including road features and obstacles and pedestrians. As in~\cite{bertozzi1998gold,choi2012environment,alessandretti2007vehicle,neven2018towards,innocenti2017imitation,chen2018efficient} authors presented an assist system in which they utilize machine vision techniques to recognize road lanes and signs. These progressive image processing methods infer lane data from forward-facing cameras mounted at the front of the vehicle~\cite{bertozzi1998gold,choi2012environment,alessandretti2007vehicle}. Some of the advanced lane finding algorithms have been developed using deep learning and neural network approaches~\cite{neven2018towards,innocenti2017imitation,chen2018efficient}. Some other procedures used for monitoring the consciousness and emotional status of the driver are momentous for the safety and comfort of driving. Nowadays, real-time non-obtrusive monitoring systems have been developed, which explore the driver's emotional states by considering facial expressions of them~\cite{fridman2018human,fridman2016owl,hoffman2000distinct,wisniewska2014robust,varma2012accident,smith2000monitoring}.

Given the nature of the safety and the fact that in previous studies the efficiency of presented methods for diagnosing the safety of car travels has been observed, hence we propose an integrated vehicle safety system, which is a compilation of the aforementioned approaches. This system can prove beneficial in terms of increasing the safety factor and driving safety and in turn, reducing crashes, casualties and the damage caused by accidents.

\section{Architecture}\label{sec:Architecture}
A system we have built is composed of a variety of subsystems, which utilize the capabilities of the machine vision and factory-installed sensors information. The following, we describe the parts and implementation stages of the system in details.

\subsection{Driving Scene Perception}\label{sec:Driving-Scene-Perception}
As we steer a vehicle, we are deciding where to go by using our eyes. The road lanes are indicated by lines on the road, which work as stable references for where to drive the car. Intuitively, one of the first things we need to do in developing a self-driving car is to identify road lane-lines using an efficient algorithm. Here is a robust approach for driving scene perception that uses trained segmentation neural network for recognizing driving safe area and extracting road along with a lane detection algorithm to deal with the curvature of the road lanes, worn lane markings, emerging/ending lane-lines, merging, splitting lanes, and lane changes.

To identify lane-lines in a video that is recorded during car driving on the road, we need a machine vision method that performs detection and annotation tasks on every frame of the video in order to generate an appropriate annotated video. The method has a processing pipeline scheme that encompasses preliminary tasks like camera calibration and perspective measurement and later stages such as distortion correction, gradient, perspective transform, processing semantic segmentation output of the deep network and lane-line detection.

The lane-line finding and localizing algorithm must be effective for real-time detecting and tracking, and has an efficient performance for different atmospheric conditions, light conditions, road curvatures, and also for other vehicles, which are in road traffic. Here, we propose an approach relied on advanced machine vision techniques to distinguish road lanes from dash-mounted camera video and detect obstacles in the car's surroundings from both of front- and rear-camera. We utilize advanced computer vision methods to compute the curvature of the road, identify lanes, and also locate the vehicle in safe driving zone. At a glance. we pursued this process into three stages, in the first stage, we calibrate a front, rear, and top cameras with correct distortion of each frame of input video and create a more suitable image for subsequent processing. In the next stage, we use a Deep Convolutional Encoder-Decoder Network that has an architecture inspired by ENet~\cite{paszke2016enet} and SegNet~\cite{badrinarayanan2015segnet} to determine potential locations of the lane-lines in the image from full input resolution feature maps for pixel-wise classification. \begin{figure}
	\centering
	\includegraphics[width=0.99\textwidth]{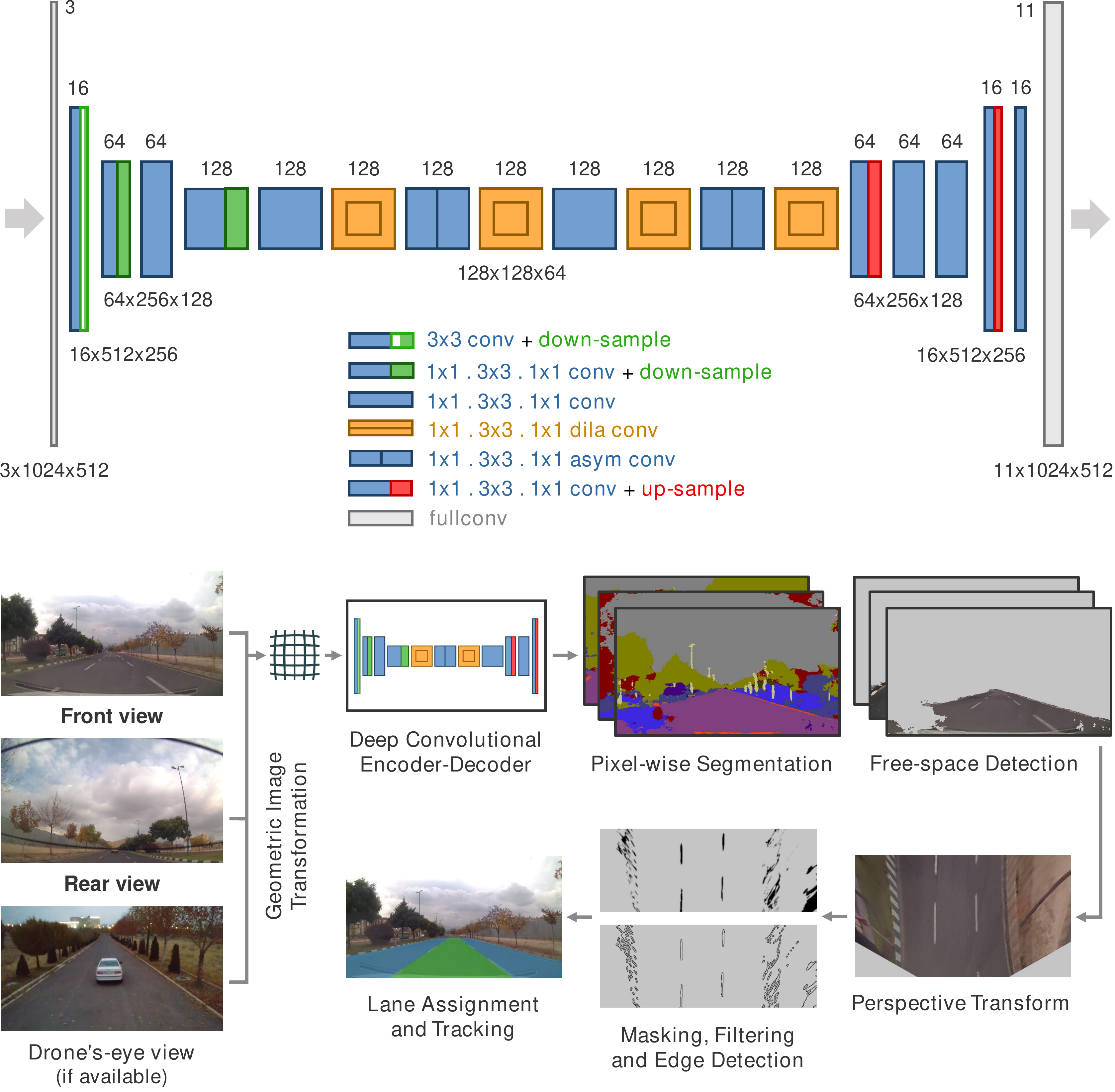}
	\caption{The overall scene understanding pipeline along with architecture of the Convolutional Encoder-Decoder Network model for scene segmentation is shown in terms of layers of convolutional networks. Each block shows different types of convolution operations (normal, full, dilated, and asymmetric). The pipeline includes geometric transformation, encoder-decoder network, free-space detection, perspective transform, masking, filtering, edge detection, lane assignment, and tracking respectively.}
	\label{fig:vision-system}
\end{figure} In the third stage we synthesize the lane mask information with prior frame information for computing the final lane-lines and identify the main vehicle's route, free-space and lane direction. This stage would be done to discard noisy effects, apply a perspective transform on the image, and track assigned lane and path (as shown in \cref{fig:vision-system}).

For this pipeline, what steps are needed to do to get a better scene understanding that is to say: first, a new frame of the video is read and then undistorted by using precomputed camera distortion matrices based on our camera's intrinsic, and extrinsic parameters, which is known as undistort image.

At second stage, we propose a deep neural network with basic encoder-decoder architecture computational unit, consisting of 17 layers, and one dimensional convolutions with small convolutional operations. Hence, training and testing are accelerated and facilitated because of lower dimensional and small convolution operations. This model leverages various types of convolution operations that are consist of regular, asymmetric, and dilated. This diversity lessens the computational load by changing dimensions of 5 × 5 convolutions in a layer into two layers with 5 × 1 and 1 × 5 convolutions~\cite{szegedy2016rethinking} and leads to fine-tuning the receptive field by the dilated convolutions application. The architecture of the encoder is similar to vanilla CNN, which includes several convolution layers with max-pooling. The encoder layers carry out feature extraction and pixel-wise classification of the down-sampled image. Somewhere else, the layers of the decoder do up-sampling after each convolutional layer for offsetting the encoder down-sampling effects and making an output with a size as same as the input. The beginning layer implements subsampling to diminish the computational load. The architecture as shown in \cref{fig:vision-system} consists of 10 convolutional layers alongside max-pooling for the encoder, 5 convolutional layers in parallel with up-sampling belong to the decoder, and a conclusive 1 × 1 convolutional layer to combine the penultimate layer outputs. All the convolution operations are either 3 × 3 or 5 × 5, whereas 5 × 5 convolutions are asymmetric, that is to say, they are performed separately as 5 × 1 and 1 × 5 convolutions to lessen the computational load. Besides, some layers use dilated convolution to increment the effective receptive field of the associated layer. Therefore, this helps with growing faster the encoder receptive field without using down-sampling. Such model is highly efficient insomuch as all convolutions are either 3 × 3 or by 5 × 5 and collateral, in contrast to sequential, integration with max-pooling potentially retains inherent details of the environmental features.

The last stage is to compute lanes. Different lane calculations would be implemented for the first frame and subsequent frames. In the initial of this stage, we apply the perspective transform in which has given bird's eye view of the road that makes to discard any irrelevant information about the background from the warped image. In the next step, once we provide the perspective transform, next, we put on color masks to recognize yellow and white pixels in the image. For final step, besides the color masks, for detecting edges we apply some filters. We use the filters on L and S channels of the image since the filters made robust the color and lighting variations. Then, we merge candidate lane pixels from color masks, filters, and pixel-wise classification map to get potential lane regions. In the first frame, the lanes are computed and determined by computer vision methods. But, in the later frames, we tracked the location of the lane-lines from previous frame. This approach significantly reduces the computation time of the algorithm. Next, we introduced additional steps to ensure some errors which might be occurred due to incorrectly detected lanes that would be removed. Last, the coefficients of the polynomial fit are used to compute curvatures of the lanes and relative location of the vehicle on the road lanes.

Ultimately, we gather all of the output results for three stages to determine the vehicle position on the road and detect free-space around the car for having a subtle defensive driving system.

\subsection{Driver State Detection}\label{sec:Driver-State-Detection}
Due to having a safe smart car, we should monitor the driver's behavior. An important component of the driver's behavior corresponds to eye-gaze tracking. Intuitively, the driver's allocation of visual attention away from the road is the momentous cause in increasing the hazards of driving. We can determine the status of the driver with their eye-gaze tracking and blink rate for detecting drowsiness and/or distraction. For monocular gaze estimation, we generally do the pupils locating and determine the inner and outer eye corners in driver's head image. \begin{figure}
	\centering
	\includegraphics[width=0.98\textwidth]{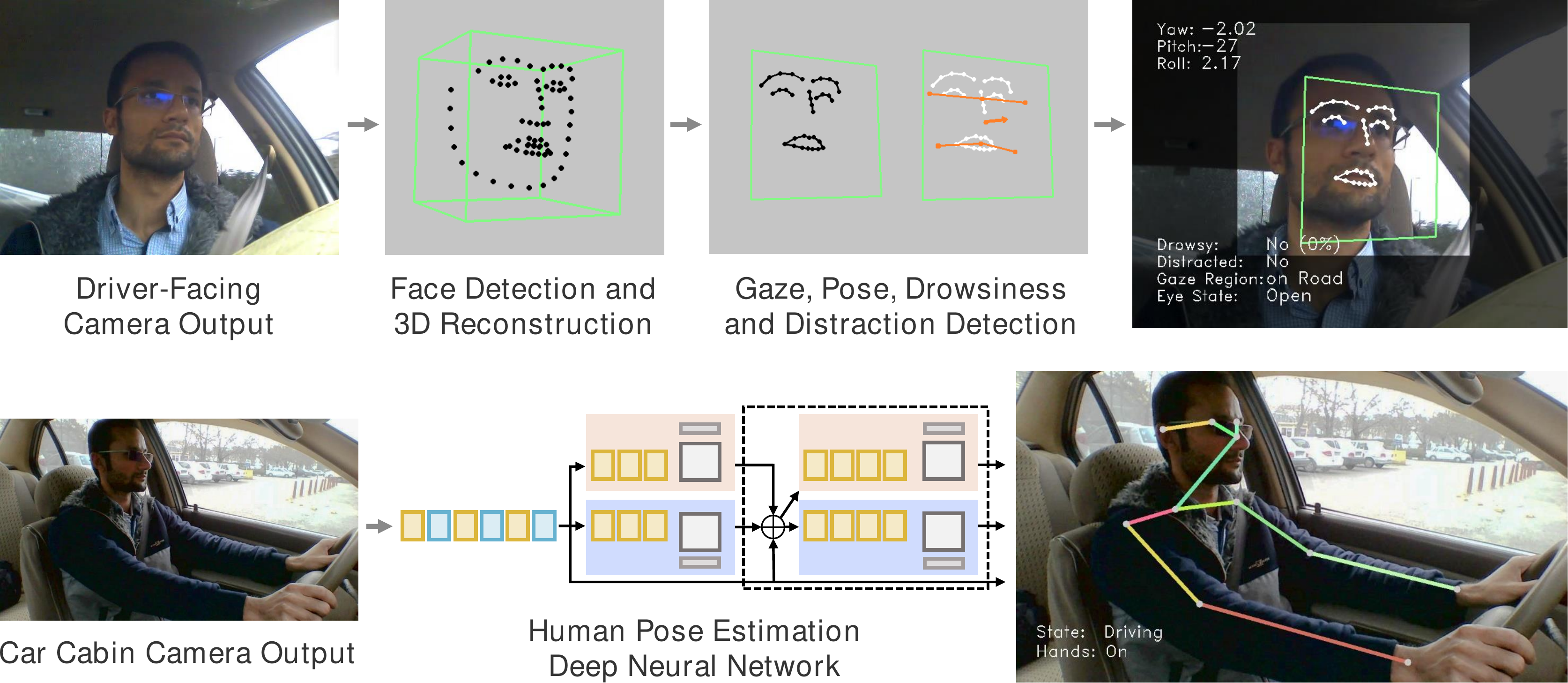}
	\caption{Driver gaze, head pose, drowsiness and distraction detection implemented in real-time for low-illumination example (top row). The computed yaw, pitch, and roll are displayed on the top left and details of the predicted state are illustrated on the bottom left. The real-time model for driver body-foot keypoints estimation on car cabin camera RGB output (bottom row), which is represent by human skeleton including head, wrist, elbow, and shoulder by color lines.}
	\label{fig:driver-monitoring}
\end{figure} Therefore, the eye corners would be as important as pupils and likely detecting them is more difficult rather than pupils. We describe how to extract the eye corners, eye region, and head pose and then utilize to estimate the gaze. The eye-gaze can be estimated using a geometric head model. If an estimate of the head pose is available, a more refined geometric model can be used and a more accurate gaze estimate is obtained. Of the locations of pupils, inner eye corner, outer eye corner, and head pose, the estimation of the eye corners is harder than other. Once the eye corners have been located, then, locating the pupils is done easily. 

In recent years, there have been done a lot of work in face identification. A novel method is proposed in~\cite{kazemi2014one} shows that face alignment can be solved with a cascade of learnt regression functions, which be able to localize the facial landmarks when initialized with the mean face pose. In the algorithm, each regression function in the cascade meticulously assesses the shape from an initial approximation and the intensities of a sparse set of pixels indexed relevant to the initial assessment. We trained our face detector as a same approach in ~\cite{kazemi2014one} by using a training set that is based on iBUG 300-W dataset, which used to learn the cascade. We determine the head pose by leveraging the proposed algorithm that estimated in a similar manner. At first, the algorithm detects and tracks a collection of anatomical feature points such as eye corners, nose, pupils, and mouth and then utilizes a geometric model to compute the head pose (as illustrated in \cref{fig:driver-monitoring}).

The steps in the driver state detection pipeline are: face detection, face alignment, 3D reconstruction, and fatigue/distraction detection. First, step for face detector we use a Histogram of Oriented Gradients (HOG) along with a SVM classifier. In this step, a false detect can be costly in the single face and multiple faces case. For the single face case, the error leads to an incorrect gaze region prediction. In the multiple faces case, the video frame would be decreasing on consideration, which reduces true decision rate at the system. Then, we perform face alignment on a 56-point subset from the 68-point Multi-PIE facial landmark markup used in the dataset. These landmarks include parts of the nose, upper edge of the eyebrows, outer and inner lips, jawline, and exclude all parts in and around the eye. Next, they would be mapped to a 3D model of the head. The resulting 3D-2D points correspondence can be used to compute the orientation of the head. This is categorized under geometric methods in~\cite{murphy2009head}. The yaw, pitch, and roll of the head can be used as features for a gaze region estimation. By using these steps, our system can indicate a gaze region recognition for each image fed into the pipeline. Given fact that the driver spends more than 90\% of their time looking forward at the road. We used this fact for normalizing facial features spot to the face bounding box, which corresponds to the road gaze region. In this step, we do not need calibration and just normalize the facial features based on eyes and nose bounding boxes only for the running frame. Eyes and nose bounding boxes are empirically found to be the most robust normalizing region. We should consider the fact that the big disorderliness in the face alignment step is correlated with the features of the jawline, the eyebrows, and the mouth.

The detected points are used to recognize eye closes and blinks. According to head pose in 3D space, we can track eye-gaze and diagnose either the driver is looking forward to the road or not. Thus, we will be able to indicate fatigue or distraction. Also, we leverage a deep neural network to perform a driver pose estimation for detecting the position and 3D orientation from major parts/joints of the body-foot keypoints (i.e. wrist, elbow, and shoulder), which is represented by human skeleton. By using this model, we can identify the status of the driver's hands and how it is positioned.

In this section, we characterized how we are able to utilize these algorithms to make a gaze assessment system that derives a desirable precision from the fact that we would be localized the corners of the eye and head pose by using the face entire appearance, rather than by just exploring a few solitary points of them.

\subsection{In-vehicle Communication Device}\label{sec:In-vehicle-Communication-Device}
\begin{figure}
	\centering
	\includegraphics[width=0.98\textwidth]{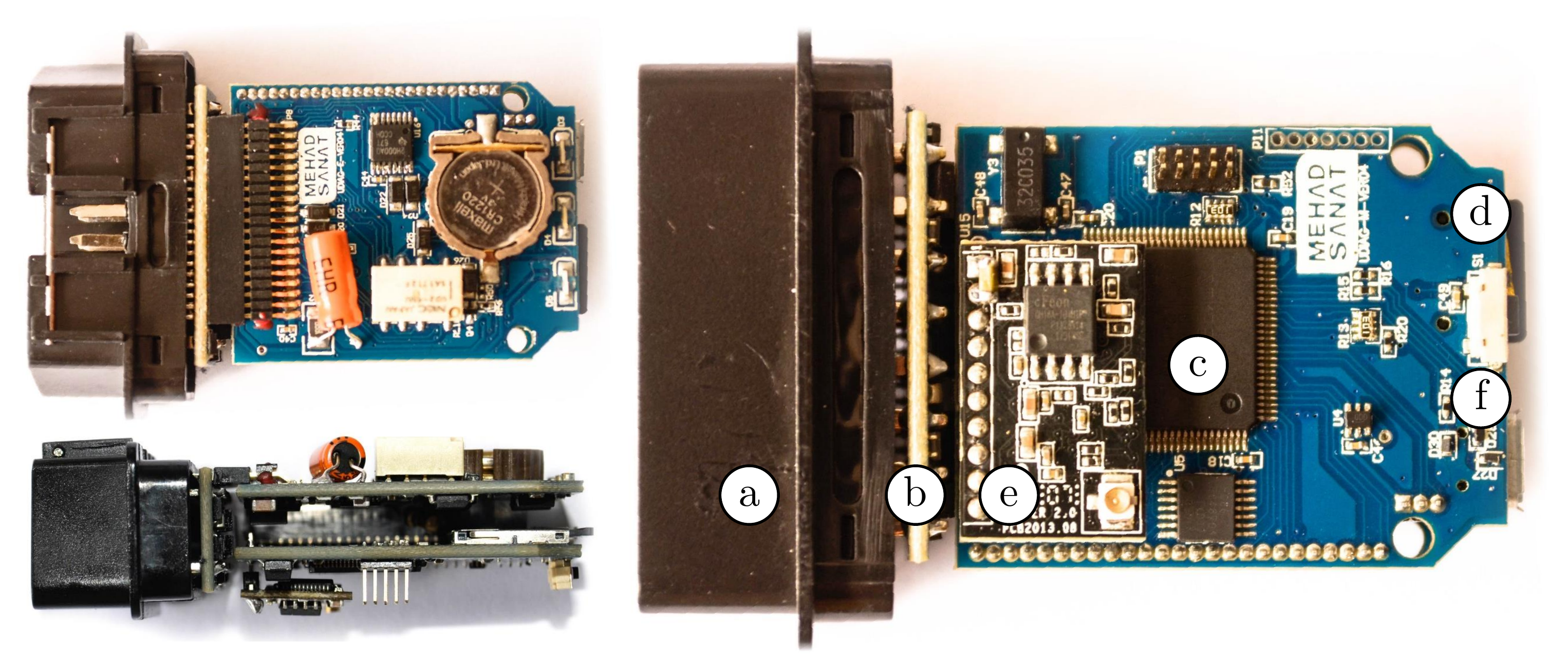}
	\caption{The top, bottom, and left view of the Universal Vehicle Diagnostic Tool (known as UDIAG) that connects to vehicle diagnostic port and establishes communications with the in-vehicle network. The vehicle network interface (a), power supply (b), processing unit (c), data storage (d), wireless adapter (e) and Micro USB socket (f) are shown in the figure.}
	\label{fig:UDIAGh-m}
\end{figure}

\begin{figure}
	\centering
	\includegraphics[width=0.98\textwidth]{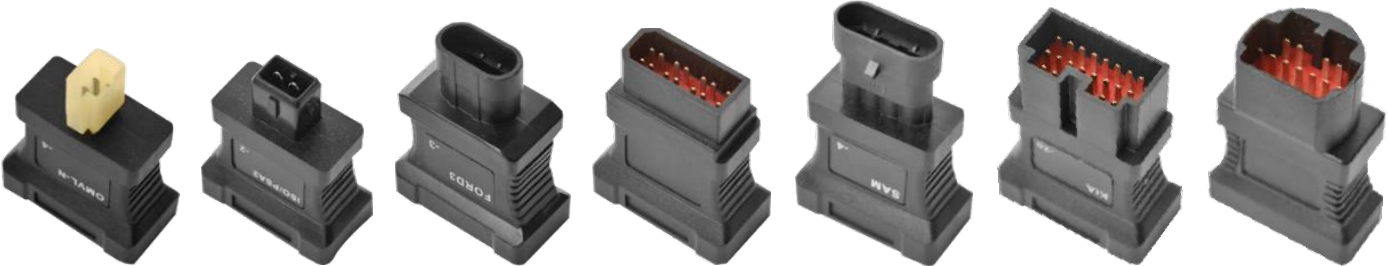}
	\caption{UDIAG external interfaces with other types of connector instead of OBD-II standard connector for communicating with various vehicles.}
	\label{fig:external-interfaces}
\end{figure}

One of the main ability of an active safety system is reliability and real-time communicating with the vehicle. In order to achieve more safety in driving with existing vehicles, we need to robust communicating with the vehicle system. For this reason, Universal Vehicle Diagnostic Tool (known as UDIAG) is developed as shown in \cref{fig:UDIAGh-m}, it is able to communicate with several types of vehicle internal communication network protocols. UDIAG connects to vehicle system via OBD-II standard connector of the vehicle directly, also it can connect to other types of connector via an external interface (\cref{fig:external-interfaces}), and negotiates with Electronic Control Units (ECUs) of the in-vehicle network according to own database. UDIAG translates data of the network into the useful and pure information such as parameter and fault codes of the vehicle and sends information via WiFi to other parts of safety system. Also, this platform injects command of safety system into in-vehicle network and saves a log of the network on own storage.

UDIAG consists of five main parts: power supply, processor, in-vehicle network interface, storage, and a wireless interface (shown in \cref{fig:UDIAGh-m}). The power supply can support both 12v and 24v vehicles. UDIAG has an ARM cortex M4 (STMF407VGT) processor, the vehicle network interface supports KWP2000, ISO 9141, J1850 and CAN~\cite{corrigan2008introduction} physical layer. For storage it uses high-speed microSD card and utilizes WiFi-UART Bridge and USB for communicating with other parts of the safety system.

We leverage the UDIAG to receive information and gather data from vehicle control units, car systems and surrounding sensors along with mounted cameras, and then process and integrate them by our system, which ultimately leads to the issuance of appropriate commands (e.g. alerting the driver to drowsiness or sudden lane changes) in various conditions.

\section{Implementation Details}\label{sec:Implementation-Details}
In order to get a safety auto-steering vehicle without the use of specific and complex infrastructures, we need to design a system that has a thorough perception of the environment and car surroundings (i.e. the road, pedestrians, other vehicles, and obstacles) at least as much as a safety threshold. Therefore in our system implementation due to take an affordable project completion, we used only passive sensors, cameras, factory-installed in-vehicle sensors, low-cost device, and an ordinary laptop in the vehicle, which allow our proposed system to be easily implemented and exploited with low operational costs. The architecture of the system which is installed and tested on the FARAZ vehicle is based on an Intel Core i5 processor along with four cameras which are consisting two wide-angle high-definition (HD) cameras, a night vision camera and webcam, and also a Universal Vehicle Diagnostic Tool called UDIAG. Two HD cameras are mounted at close to the top center of the windshield and rear window that are used to take video from front perspective to detect the road and lane-lines and rear perspective to detect other vehicles and obstacles in car's surroundings. One camera is mounted on dashboard to supervise the face of the driver for detecting fatigue, drowsiness and/or driver's distraction. Another webcam is mounted on the headliner close to the top center of the windshield that used for driver body pose monitoring. C++ programming language, OpenCV (Open Source Computer Vision Library), and FreeRTOS (Free Real Time Operating System) have been used for a complete implementation of the system.

The process of the system is such that all data are collected from sensors and commands are received from the user interface, which can be entered through the system's control panel namely graphical user interface or keyboard. After analyzing input data, the system leverages extracted information to decide on the measures to perform regarding suited warnings and driving strategies. Also, for debugging purposes, a visual output would be supplied to the user and intermediate results are logged. FARAZ, shown in \cref{fig:instrumented-vehicle} is an experimental semi-autonomous vehicle equipped with a vision system and a supervised steering capability. It is able to determine its position with respect to road lane-lines, compute the road geometry, detect generic obstacles on the trajectory, and assign the vehicle to a lane and maimtain the optimal path. The system is designed as a safety enhancement unit. In particular, it is able to supervise the driver behavior and issue both optic and acoustic warnings. It issues a proper command/alert at speeding, sudden road lane changes, encountering an obstacle on car's route, when approaching to a vehicle's rear or vice versa and the possibility of rear-end collision, sudden crash around the car, when to drive slower than traffic, and even the need to fix the automobile using the information, which is acquired from car systems.

We can adjust the system to steer the car in two different modes: a \textit{manual mode} that the system monitors and logs the driver's activity, and alerts of hazard cases to driver with acoustic and optical warnings. Data logging while driving in the system includes important data such as speed, lane detection and changes, user interventions, and commands. \textit{Semi-automated mode} that in addition to warning and log capabilities, it also sends some controlling commands to car systems and even is able to take control of the vehicle when a dangerous situation is detected and also we equipped the  car FARAZ with emergency devices that can be activated manually in case of system failures. Further, for future work, we will add an automated mode in the system that leads to full control on the vehicle.

The FARAZ car being used in our tests has eight ECUs for various tasks: Central Communication Node (CCN) in the dashboard to manage the central locking and alarm system and to communicate with the body modules, the lighting system, and read the status of the various switches, Door Control Node (DCN) for controlling door actuators and vehicle mirrors, Front Node (FN) on front of the vehicle to control the alternator, cooler compressor, horn and lights set, car alarms, and front actuators, Instrument Cluster Node (ICN) to control various front-end amps, Rear Node (RN) in the rear luggage compartment for rear-end car sensors and lights, Anti-lock Braking System (ABS) for the management of brakes and vehicle wheels, Airbag Control Unit (ACU) for airbags and related actuators, and Engine Management System (EMS), which is responsible for driving the engine vehicle and sending control commands. The status information and values of the actuators and car sensors associated with these modules are read from the internal vehicle network and sent to the integrated safety system for decision making.

Values or status of vehicle speed, engine speed, engine status, throttle position, throttle angle, acceleration pedal angle, battery voltage, mileage, gearbox ratio and engine configuration from EMS, the speed of each wheel individually from ABS, the relevant information for each airbag from ACU, information on all switches (e.g. the wash pump, wiper, air condition, screen heater) inside the vehicle and in the car bonnet, information on the status of all car lamps (such as main, dipped, fog, side, hazard), hand break and brake pedal status, shock sensor status, seat belt status, gasoline level, the status of each car doors and mirrors, outdoor and indoor temperature, brake oil level, oil pressure, cruise control and target velocity of cruise control (if available) from CCN, DCN, FN, RN, and ICN nodes, and also the status of the central locking (Locked/Unlocked) and key position are obtained from the Immobilizer indirectly. Our device can also send appropriate commands for each of the actuators associated with each of the different modules according to the decision-making conditions.

Our vehicle has had decentralized road tests within a month in Zanjan. Each part of the system, as described in the previous section, was tested on the training data and validated before the final test on the vehicle. Then all of them have been put together to check the functionality of the system. The initial tests were conducted to check the overall performance of the vehicle along with a driver at all times of the test, on the campus paths and urban roads in a controlled environment of possible incidents (pedestrian crossings, car accidents, etc.). These tests were carried out at different times of the day and night with a cover distance of 100km in normal climate conditions. In the future, these tests would be carried out on a long-term schedule. It also seeks to further implement this system on a commercial vehicle with more ECUs and more environmental sensors to add fully autonomous system capabilities.

\section{Conclusions and Future Work}\label{sec:Conclusions-and-Future-Work}
This paper describes a developed safety system for a human-centered semi-autonomous vehicle designed to detect mistakes in driver behavior where system's perception pipeline for the driving function faces an edge-case, which driver might be struggling but is not conscious of it, and then the system offers a proper alert or even issues an appropriate command. Our system applies a deep convolutional encoder-decoder network model to serve as a secondary appliance along with a vision system installed on our vehicle for the driving scene perception. In addition, we leveraged a universal in-vehicle network device to control the entire system, establish communication with each component of the system, and check all parts of the system to be properly enabled. We show that the proposed system is able to act as an effective supervisor with issuing proper steering commands and proportionate measures during driving. Our system is capable of detecting driver errors in less than 2 seconds using the cameras embedded in the car cabin. Thanks to the UDIAG, the system is also able to read and log all the information about the car's ECUs. Collected data is used for a more subtle decision-making process in the system, and using this information in the future, we can achieve a better end-to-end model for autonomous driving.

For future work, we schedule to add an ability to monitor vehicle status on the road through drone’s-eye view which has an auto-guidance system, and eke to examine and evaluate the system on today's modern vehicles with advanced navigation systems in different weather conditions.

{\subsection*{Acknowledgments}}
This project was in part supported by a grant from Mehad Sanat Incorporation and Institute for Research in Fundamental Sciences (IPM).
The research was partially supported by OP RDE project No. CZ.02.2.69/0.0/0.0/16\_027/0008495, International Mobility of Researchers at Charles University.

Our team gratefully acknowledges researchers and professional engineers from Mehad Sanat Incorporation for the automotive technical consultant and offering hardware equipment.
\\

\renewcommand{\bibsection}{\section*{References}}
\bibliographystyle{splncs04}
\begingroup
  \ifluatex
  \else
    \microtypecontext{expansion=sloppy}
  \fi
  \small % ensure correct font size for the bibliography
  \bibliography{An-Intelligent-Safety-System-for-Semi-Autonomous-Vehicle}
\endgroup

\end{document}